\documentclass[10pt,twocolumn]{article}

\usepackage[utf8]{inputenc}
\usepackage[T1]{fontenc}
\usepackage{lmodern}
\usepackage{microtype}

\usepackage{graphicx}
\usepackage{subcaption}
\usepackage{booktabs}
\usepackage{multirow}
\usepackage{tabularx}
\usepackage{amsmath, amssymb, amsfonts}
\usepackage{xcolor}
\usepackage{listings}
\usepackage{mathtools}
\usepackage{bm}
\usepackage{siunitx}
\usepackage{algorithm}
\usepackage{algorithmic}

\usepackage[hidelinks]{hyperref}
\usepackage[nameinlink]{cleveref}

\usepackage{geometry}
\newcolumntype{L}[1]{>{\raggedright\arraybackslash}p{#1}}
\newcolumntype{C}[1]{>{\centering\arraybackslash}p{#1}}
\newcolumntype{Y}{>{\raggedright\arraybackslash}X}

\usepackage[numbers]{natbib}

\usepackage{authblk}

\title{ChronoDreamer: Action-Conditioned World Model as an Online Simulator for Robotic Planning}

\author[1]{Zhenhao Zhou, Dan Negrut}
\affil[1]{University of Wisconsin--Madison}
\affil[ ]{\texttt{zzhou292@wisc.edu, negrut@wisc.edu}}
\date{} 

\begin{document}
\maketitle

\begin{abstract}
We present ChronoDreamer, an action-conditioned world model for contact-rich robotic manipulation. Given a history of egocentric RGB frames, contact maps, actions, and joint states, ChronoDreamer predicts future video frames, contact distributions, and joint angles via a spatial-temporal transformer trained with MaskGIT-style masked prediction. Contact is encoded as depth-weighted Gaussian splat images that render 3D forces into a camera-aligned format suitable for vision backbones. At inference, predicted rollouts are evaluated by a vision-language model that reasons about collision likelihood, enabling rejection sampling of unsafe actions before execution. We train and evaluate on DreamerBench, a simulation dataset generated with Project Chrono that provides synchronized RGB, contact splat, proprioception, and physics annotations across rigid and deformable object scenarios. Qualitative results demonstrate that the model preserves spatial coherence during non-contact motion and generates plausible contact predictions, while the LLM-based judge distinguishes collision from non-collision trajectories.
\end{abstract}

\section{Introduction}

Robots operating in contact-rich environments require trajectory planning that respects both task objectives and collision constraints. Classical approaches rely on explicit geometric models and physics simulators, but high-fidelity simulation is often too slow for online replanning, and sim-to-real gaps persist even with careful calibration. Learned world models offer a complementary path: by predicting future states conditioned on actions, they enable planning by imagination at speeds compatible with real-time control. However, most video prediction models focus on visual plausibility and neglect the physical quantities---contact forces, friction modes, joint states---that determine whether a trajectory is safe.

This work addresses the gap between video generation and contact-aware planning. We introduce ChronoDreamer, an action-conditioned world model that jointly predicts future RGB frames, contact maps, and joint angles. The model operates on discrete visual tokens from a pretrained encoder and uses a spatial-temporal transformer with factorized attention to maintain tractable complexity over long horizons. Contact is represented as a camera-aligned image via depth-weighted Gaussian splats, encoding force magnitude and direction in a form directly consumable by vision backbones. At inference time, predicted rollouts are passed to a vision-language model that reasons about collision likelihood, enabling rejection sampling of unsafe actions before execution.

\paragraph{Contributions.}
\begin{itemize}
  \item A contact encoding scheme that renders 3D contact forces as depth-weighted Gaussian splat images aligned with the robot's egocentric camera, providing dense contact supervision in an image-native format.
  \item ChronoDreamer, a spatial-temporal transformer world model that predicts video tokens, contact tokens, and joint angles jointly via MaskGIT-style masked prediction with factorized vocabulary.
  \item Integration of world-model rollouts with an LLM-based collision judge that reasons over predicted frames and contact maps to filter unsafe action candidates online.
  \item Evaluation on DreamerBench, a multi-scenario dataset with rigid and deformable objects, demonstrating spatial coherence preservation and contact-event prediction.
\end{itemize}

\section{Background and Motivation}
State-of-the-art text- and image-conditioned video generators now produce photorealistic textures and intricate geometry, yet they still violate basic mechanics: objects interpenetrate, trajectories defy gravity, and energy appears or vanishes.
For robotics, scientific visualization, or simulation-aware media, these failures are structural rather than aesthetic, and they motivate returning to the world-model viewpoint in which a generator maintains an internal state and transition dynamics coherent enough to support imagination and planning.
Classic systems such as World Models, PlaNet, and Dreamer---and large-scale successors like Genie---demonstrated that long-horizon behavior hinges on latent dynamics that stay faithful to first principles~\cite{ha2018worldmodels}.
Reframing modern video generation through this lens suggests that perceptual fidelity must be paired with state updates governed by constraints resembling Newtonian, and more broadly physical, laws.

One route to lawful dynamics is to make physically plausible motion the default attractor via architectural bias.
PhyDNet and PhyLoNet introduce dual pathways that disentangle a PDE-inspired ``physics'' cell from a residual appearance branch; the former evolves smooth, conservative dynamics, while the latter handles nuisance visual factors~\cite{LeGuen2020PhyDNet}.
Even though these designs predate diffusion models, the lesson transfers: inserting lightweight, physics-structured latent updates between denoising steps or constraining attention flows to respect locality regularizes temporal evolution, cuts down implausible transitions, and keeps the latent state easier to plan in---much like traditional world-model agents that rely on carefully structured transition models instead of unconstrained recurrence.

Another strand delegates state evolution to a mechanistic simulator and reserves the generator for photorealistic rendering.
PhysGen exemplifies a staged pipeline in which geometry and material parameters are inferred from a single image and user command, a rigid-body simulator propagates contact-rich dynamics, and a diffusion renderer produces temporally consistent imagery guided by the simulated motion~\cite{Liu2024PhysGen}.
Variants insert deformable or continuum solvers, or periodically project denoising intermediates back onto the simulator’s feasible set, tightening constraints while preserving generative diversity.
From the world-model perspective, these simulator-in-the-loop systems externally scaffold the transition model: a trusted dynamical engine maintains or corrects the latent state, whereas the generator focuses on appearance.
Design choices revolve around the governing simulator class, how tightly it couples to the denoiser, and how physical parameters are estimated or adapted over time, but even simple staged interfaces yield large reductions in violations while preserving user control.

A complementary approach relies on multimodal language and vision-language models to supply symbolic reasoning about forces, contacts, and constraints before or during sampling.
DiffPhy parses a prompt to extract entities and qualitative dynamics, then guides a diffusion model to honor that physical ``plan''; VideoREPA distills relational understanding from recognition models into text-to-video generators so that high-level commonsense physics is baked into the latent transitions without embedding a full simulator~\cite{zhang2025thinkdiffuse}.
Viewed through world modeling, the reasoner acts as a high-level dynamics prior that proposes coarse motion plans which the generator follows, yielding more globally plausible rollouts while retaining the coverage of large video backbones.
Because symbolic reasoning can hallucinate, practical systems pair it with lightweight validators or occasional simulator hints, but the amortized planning cost makes this path attractive when prompts describe qualitative effects that are difficult to capture with explicit simulation alone.

Finally, physics can be enforced through training-free guidance signals that nudge the denoiser away from violations without modifying the base model.
Recent work constructs counterfactual prompts that deliberately break physics and uses synchronized, decoupled guidance to push samples away from those modes, improving gravity and contact behavior without finetuning~\cite{hao2025implausibility}.
Related frame-level control techniques inject depth, sketch, or keyframe cues during denoising; when those cues are replaced by physics-violation critics such as penetration detectors or anti-gravity classifiers, the same machinery becomes a soft controller on the sampler’s transition dynamics~\cite{jang2025frameguidance}.
By training lightweight critics or distilling simulator judgments into differentiable surrogates, one can retrofit powerful pretrained generators with practical physics correctors, combining the engineering pragmatism of plug-and-play guidance with the rigor of simulator or reasoning-based constraints.

\section{Method}
\label{sec:method}

\subsection{Contact Encoding via Depth-Weighted Gaussian Splats}
\label{subsec:dreamerbench:contact-encoding}

A central design choice in DreamerBench is to represent contact as a
camera-centric image rather than as a sparse list of contact points.
Existing work explores several alternatives. Neural Contact
Fields represent contact as a continuous probability field over 3D
space conditioned on tactile inputs, enabling estimation of complex
extrinsic contact patches from visuotactile signals~\cite{higuera2022ncf}.
Vision-only approaches such as Im2Contact recover dense contact maps
from RGB observations by learning a pixel-wise contact likelihood
without explicit access to forces or tactile
measurements~\cite{kim2023im2contact}. Large-scale visuotactile
datasets including FreeTacMan encode contact as image-like tactile
fields synchronized with external views for contact-rich
manipulation~\cite{wu2025freetacman}. More recently, DyTact models
dynamic contacts using 2D Gaussian surfels attached to articulated
meshes, rasterized into time-varying contact
fields~\cite{cong2025dytact}. These lines of work suggest that
contact fields expressed in an image-like domain are a convenient
interface for modern vision backbones and world models.

In DreamerBench, each simulation step produces a \emph{contact splat}
image that aggregates all active contacts into a single RGB frame aligned
with a reference camera. For each contact $i$, we log a 3D contact
position $\mathbf{p}_i \in \mathbb{R}^3$ and a contact force
$\mathbf{f}_i \in \mathbb{R}^3$. Given a camera with world-frame
position $\mathbf{c} \in \mathbb{R}^3$ and rotation
$\mathbf{R}_{cw} \in \mathrm{SO}(3)$ (rotation from world to camera), the
camera-frame coordinates of the contact point are
\begin{equation}
  \mathbf{x}_i
  = \mathbf{R}_{cw} (\mathbf{p}_i - \mathbf{c})
  = \begin{bmatrix} X_i \\ Y_i \\ Z_i \end{bmatrix}.
\end{equation}
We interpret $X_i$ as depth along the camera forward axis (with $X_i>0$
in front of the camera) and apply near-plane clipping by discarding
contacts with $X_i < X_{\min}$ or clipping line segments to the plane
$X = X_{\min}$ to avoid numerical instabilities in the projection.

Projection to the discrete image plane uses a pinhole model with
intrinsics $(f_x,f_y,c_x,c_y)$:
\begin{equation}
  u_i = c_x - f_x \frac{Y_i}{X_i}, \qquad
  v_i = c_y - f_y \frac{Z_i}{X_i},
\end{equation}
followed by rounding to the nearest pixel center
$(\tilde{u}_i,\tilde{v}_i) \in \mathbb{Z}^2$. To encode force
\emph{direction} in the image plane, the implementation also projects a
second point displaced along the force vector,
\begin{equation}
  \mathbf{x}_i^{\text{end}}
  = \mathbf{R}_{cw}\bigl(\mathbf{p}_i + s_i \mathbf{f}_i - \mathbf{c}\bigr),
\end{equation}
with a scale $s_i$ chosen as an affine or quantile-based function of the
force magnitude $\|\mathbf{f}_i\|$. The displacement between the two
projections in pixel space,
\begin{equation}
  \Delta \mathbf{d}_i
  = \begin{bmatrix} \Delta u_i \\ \Delta v_i \end{bmatrix}
  = \begin{bmatrix} u_i^{\text{end}} - u_i \\ v_i^{\text{end}} - v_i \end{bmatrix},
\end{equation}
defines a 2D direction vector. After normalization
$\hat{\mathbf{d}}_i = \Delta \mathbf{d}_i / \|\Delta \mathbf{d}_i\|$,
the two components are mapped from $[-1,1]$ to $[0,1]$ and stored in the
green and blue channels, respectively:
\begin{equation}
  G_i = \frac{\hat{d}_{i,x} + 1}{2}, \qquad
  B_i = \frac{\hat{d}_{i,y} + 1}{2}.
\end{equation}
The red channel encodes a clipped and normalized force magnitude,
\begin{equation}
  m_i = \|\mathbf{f}_i\|, \qquad
  R_i = \frac{\min(m_i, m_{\max})}{m_{\max}} \in [0,1],
\end{equation}
so that $R_i$ saturates for very large contact forces.

Each contact is then rendered as an isotropic Gaussian kernel centered at
$(\tilde{u}_i,\tilde{v}_i)$, with a radius that depends on the
normalized magnitude $m_i/m_{\max}$. Specifically, we define a radius
$r_i$ through
\begin{equation}
  t_i = \left(\frac{m_i}{m_{\max}}\right)^\gamma, \qquad
  r_i = r_{\min} + (r_{\max} - r_{\min}) t_i,
\end{equation}
with a shaping exponent $\gamma \ge 1$ and bounds
$0 < r_{\min} \le r_{\max}$. The corresponding Gaussian kernel over
integer pixel offsets $(\delta u,\delta v)$ is
\begin{equation}
  k_i(\delta u, \delta v)
  = \exp\!\left(
      -\frac{\delta u^2 + \delta v^2}{2\sigma_i^2}
    \right),
  \quad \sigma_i \approx \frac{r_i}{3},
\end{equation}
cropped to a finite window of radius $\mathcal{O}(r_i)$.
To encourage nearer contacts to dominate farther ones, the kernel is
multiplied by a depth-dependent weight
\begin{equation}
  w_i^{\text{depth}}
  = \exp\!\left(-\frac{X_i}{\tau_{\text{depth}}}\right),
\end{equation}
with a tunable depth scale $\tau_{\text{depth}} > 0$. The total weight
for contact $i$ at pixel $(u,v)$ is then
\begin{equation}
  w_i(u,v) = w_i^{\text{depth}}\, k_i\bigl(u - \tilde{u}_i,\, v - \tilde{v}_i\bigr).
\end{equation}

Let $\mathbf{c}_i = (R_i,G_i,B_i)^\top$ denote the per-contact color
vector. The image is assembled by weighted accumulation over all
contacts, followed by a per-pixel normalization:
\begin{align}
  \tilde{\mathbf{I}}(u,v)
  &= \sum_{i} w_i(u,v)\, \mathbf{c}_i, \\
  W(u,v)
  &= \sum_{i} w_i(u,v), \\
  \mathbf{I}(u,v)
  &= 
  \begin{cases}
    \tilde{\mathbf{I}}(u,v) / W(u,v),
    & \text{if } W(u,v) > 0, \\
    \mathbf{0},
    & \text{otherwise.}
  \end{cases}
\end{align}
Here $\mathbf{I}(u,v) \in [0,1]^3$ is the final contact splat image at
the given time step. The normalized averaging behaves like a soft
z-buffer over Gaussian ``surfels'': overlapping contributions are blended
rather than summed unboundedly, and nearer contacts are given higher
effective weight through $w_i^{\text{depth}}$. The resulting image
compactly encodes contact spatial footprint, approximate pressure (via
red intensity and kernel size), and projected force direction (via
green–blue hue) in a form directly consumable by convolutional networks
or video-tokenization modules.

\subsection{World Model Architecture}

This section describes ChronoDreamer, a spatial-temporal transformer architecture for action-conditioned video prediction that operates on discrete latent representations obtained through finite-scalar quantization. The model employs a MaskGIT-style training objective for next-frame prediction and jointly predicts future video frames, contact maps, and joint angles conditioned on historical observations and action sequences.

\subsubsection{Architecture Overview}

\begin{figure}[t]
  \centering
  \includegraphics[width=0.48\textwidth]{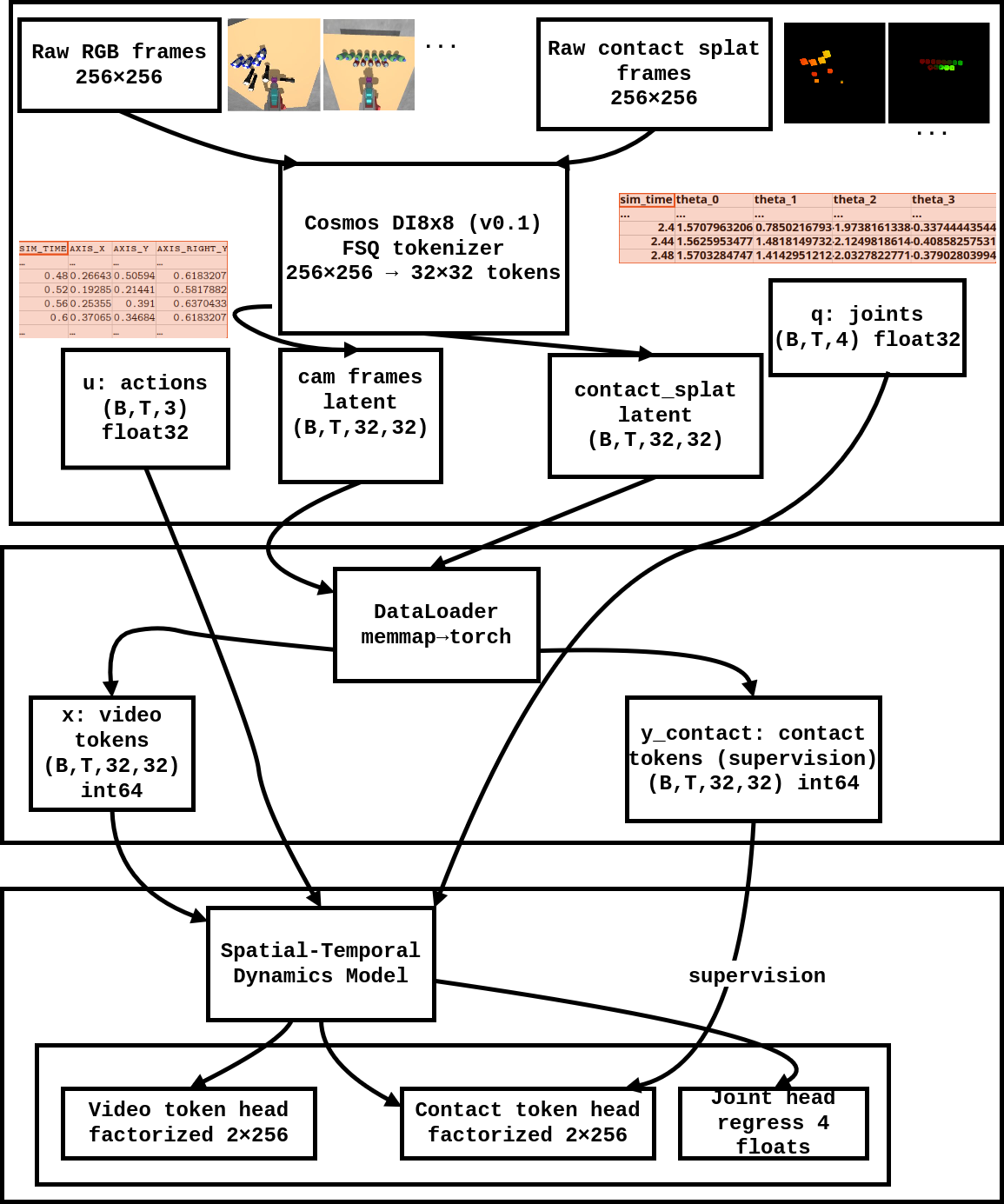}
  \caption{Data pipeline. Cosmos encoder tokenizes RGB and contact frames into $32 \times 32$ grids. The dynamics model outputs factorized video/contact logits and joint angles.}
  \label{fig:architecture-overview}
\end{figure}

Figure~\ref{fig:architecture-overview} depicts the data pipeline. Raw RGB frames and contact splat images at $256 \times 256$ resolution are tokenized by the Cosmos DI8$\times$8 encoder via finite-scalar quantization, yielding $32 \times 32$ discrete token grids per frame. These tokens, along with continuous action commands and joint angle observations, are loaded via memory-mapped I/O and passed to the spatial-temporal dynamics model, which outputs predicted video tokens, contact tokens, and regressed joint angles.

The architecture comprises four components. The \emph{video encoder} (Nvidia Cosmos Tokenizer) maps $H \times W$ RGB frames to $h \times w$ discrete token grids with vocabulary size $V = 65{,}536$ via finite-scalar quantization. The \emph{spatial-temporal transformer} consists of $L$ blocks, each containing factorized spatial and temporal self-attention followed by a feed-forward network. The \emph{MaskGIT module} performs masked token prediction with iterative unmasking governed by a cosine schedule. The \emph{video decoder} reconstructs RGB frames from quantized latents; separate linear heads project to contact map logits and joint angle outputs.

Throughout this section, we denote the batch size by $B$, temporal sequence length by $T$, input frame dimensions by $H \times W$, latent token grid dimensions by $h \times w$ (with $S = h \times w$ spatial tokens per frame), model hidden dimension by $D$, number of attention heads by $N_h$, vocabulary size by $V$, and number of transformer layers by $L$.

\subsubsection{Video Encoder: Nvidia Cosmos Tokenizer with Finite-Scalar Quantization}

\paragraph{Encoder Architecture}

The video encoder transforms input RGB frames $\mathbf{x} \in \mathbb{R}^{3 \times H \times W}$ into discrete latent tokens $\mathbf{z} \in \{0, 1, \ldots, V-1\}^{h \times w}$. The encoding process comprises two stages: continuous feature extraction and lookup-free quantization.

\paragraph{Convolutional Feature Extraction}

The encoder employs a hierarchical convolutional architecture with residual blocks. Given an input image $\mathbf{x}$, the encoder first applies an initial convolution:
\begin{equation}
    \mathbf{h}_0 = \text{Conv}_{3 \times 3}(\mathbf{x}; C_{\text{base}})
\end{equation}
where $C_{\text{base}} = 128$ is the base channel dimension.

The encoder consists of $N_{\text{blocks}} = 5$ stages with channel multipliers $(1, 1, 2, 2, 4)$. At each stage $i$, the feature map passes through $N_{\text{res}} = 2$ residual blocks:
\begin{equation}
    \mathbf{h}_i^{(j)} = \mathbf{h}_i^{(j-1)} + \mathcal{F}_{\text{res}}(\mathbf{h}_i^{(j-1)})
\end{equation}
where the residual function $\mathcal{F}_{\text{res}}$ is defined as:
\begin{multline}
    \mathcal{F}_{\text{res}}(\mathbf{h}) = \text{Conv}_{3 \times 3}\bigl(\sigma\bigl(\text{GN}\bigl(\\\text{Conv}_{3 \times 3}\bigl(\sigma\bigl(\text{GN}(\mathbf{h})\bigr)\bigr)\bigr)\bigr)\bigr)
\end{multline}
with $\text{GN}(\cdot)$ denoting group normalization with 32 groups and $\sigma(\cdot)$ the Swish activation function:
\begin{equation}
    \sigma(x) = x \cdot \text{sigmoid}(x)
\end{equation}

Spatial downsampling is performed between stages using strided convolutions:
\begin{equation}
    \mathbf{h}_{i+1}^{(0)} = \text{Conv}_{3 \times 3}^{\text{stride}=2}(\mathbf{h}_i^{(N_{\text{res}})})
\end{equation}

The final encoder output is obtained through:
\begin{equation}
    \mathbf{e} = \text{Conv}_{1 \times 1}\left(\sigma\left(\text{GN}\left(\mathbf{h}_{\text{mid}}\right)\right); C\right)
\end{equation}
where $C = 6$ is the number of continuous latent channels (before quantization) and $\mathbf{h}_{\text{mid}}$ denotes the output of additional mid-block residual layers. The encoder architecture includes a 2-level Haar wavelet transform applied to the input for initial downsampling (factor of 4).

\paragraph{Finite-Scalar Quantization (FSQ)}

Finite-Scalar Quantization (FSQ) is a non-learned discrete quantization method that maps continuous latent channels to discrete token indices. For each of the $C = 6$ continuous latent channels, FSQ independently quantizes values to a fixed set of discrete levels. The resulting token indices range over $V \approx 65{,}536$ distinct values, allowing the encoder to represent rich visual information without maintaining explicit learnable codebooks.

During decoding, given a token index, FSQ internally de-quantizes to recover the original continuous latent representation.

\paragraph{Quantization and Token Index Computation.} FSQ operates on the $C = 6$ continuous latent channels independently. Each channel is mapped to a discrete level, and the combination of all $C$ channels produces a single token index in the range $[0, V-1]$ where $V \approx 65{,}536$. The quantization is deterministic (non-learned) and fully differentiable.

\paragraph{Entropy Regularization.} To encourage codebook utilization, an entropy loss is applied:
\begin{equation}
    \mathcal{L}_{\text{entropy}} = \alpha_{\text{sample}} \cdot H_{\text{sample}} - \alpha_{\text{batch}} \cdot H_{\text{batch}}
\end{equation}
where $H_{\text{sample}}$ is the average per-sample entropy and $H_{\text{batch}}$ is the entropy of the batch-averaged token distribution. The coefficients $\alpha_{\text{sample}} = \alpha_{\text{batch}} = 1.0$ balance sample-level minimization (encouraging deterministic assignments) and batch-level maximization (encouraging uniform codebook usage).

\paragraph{Token Factorization}

For large vocabulary sizes, token embeddings are factorized to reduce parameter count. Given $V = 2^{18}$ and a factorization into $K = 2$ sub-vocabularies, each token index $z$ is decomposed:
\begin{equation}
    z = z_1 \cdot V_f + z_0
\end{equation}
where $V_f = \sqrt{V} = 512$ is the factored vocabulary size, and $z_0, z_1 \in \{0, \ldots, V_f - 1\}$.

The token embedding is computed as the sum of factored embeddings:
\begin{equation}
    \mathbf{E}(z) = \mathbf{E}_0(z_0) + \mathbf{E}_1(z_1)
\end{equation}
where $\mathbf{E}_0, \mathbf{E}_1 \in \mathbb{R}^{V_f \times D}$ are learnable embedding matrices.

\subsubsection{Input Representation and Conditioning}

\paragraph{Input Modalities}

The model receives four input modalities: \emph{history video tokens} $\mathbf{Z}_{\text{hist}} \in \{0, \ldots, V\}^{T_h \times S}$, discrete tokens for $T_h$ history frames with $S = h \times w$ spatial tokens per frame; \emph{history actions} $\mathbf{a}_{\text{hist}} \in \mathbb{R}^{T_h \times 3}$, continuous vectors encoding commanded velocities; \emph{future actions} $\mathbf{a}_{\text{fut}} \in \mathbb{R}^{T_f \times 3}$, action vectors for frames to be predicted; and \emph{history joint angles} $\boldsymbol{\theta}_{\text{hist}} \in \mathbb{R}^{T_h \times N_j}$ with $N_j = 4$ joint channels.

\paragraph{Embedding Strategy}

\textit{Video Token Embedding.} Video tokens are embedded using the factorized embedding scheme:
\begin{equation}
    \mathbf{X}_{\text{video}} = \text{FactorizedEmbed}(\mathbf{Z}) \in \mathbb{R}^{B \times T \times S \times D}
\end{equation}

A special mask token embedding $\mathbf{e}_{\text{mask}} \in \mathbb{R}^D$ is used for tokens that are masked during training:
\begin{equation}
    \mathbf{X}_{\text{video}}[i,t,s] = \begin{cases}
        \text{FactorizedEmbed}(z_{t,s}) & \text{if } z_{t,s} \neq z_{\text{mask}} \\
        \mathbf{e}_{\text{mask}} & \text{otherwise}
    \end{cases}
\end{equation}

\textit{Action Embedding.} Actions are projected to the model dimension through a linear layer followed by layer normalization:
\begin{equation}
    \mathbf{A} = \text{LayerNorm}(\mathbf{W}_a \mathbf{a} + \mathbf{b}_a) \in \mathbb{R}^{B \times T \times D}
\end{equation}
where $\mathbf{W}_a \in \mathbb{R}^{D \times 3}$ and $\mathbf{b}_a \in \mathbb{R}^D$.

\textit{Joint Angle Embedding.} Joint angles are similarly projected:
\begin{equation}
    \mathbf{J} = \text{LayerNorm}(\mathbf{W}_j \boldsymbol{\theta} + \mathbf{b}_j) \in \mathbb{R}^{B \times T \times D}
\end{equation}
where $\mathbf{W}_j \in \mathbb{R}^{D \times N_j}$.

\textit{Combined Control Token.} The action and joint angle embeddings are combined additively to form a single control token per frame:
\begin{equation}
    \mathbf{C}_t = \mathbf{A}_t + \mathbf{J}_t \in \mathbb{R}^D
\end{equation}

This combined control token is prepended to the video tokens for each frame, yielding an augmented sequence:
\begin{equation}
    \mathbf{X}_t = [\mathbf{C}_t; \mathbf{X}_{\text{video}, t}] \in \mathbb{R}^{(S+1) \times D}
\end{equation}

The full input tensor has shape $(B, T, S+1, D)$, where the first position in the spatial dimension corresponds to the control token.

\paragraph{Positional Encoding}

Learnable positional embeddings encode both temporal and spatial positions:
\begin{equation}
    \mathbf{P} \in \mathbb{R}^{1 \times T \times (S+1) \times D}
\end{equation}

The positional embedding is added to the input representation:
\begin{equation}
    \mathbf{X}' = \mathbf{X} + \mathbf{P}
\end{equation}

\subsubsection{Spatial-Temporal Transformer Decoder}

\paragraph{Architecture Overview}

\begin{figure}[t]
  \centering
  \includegraphics[width=0.48\textwidth]{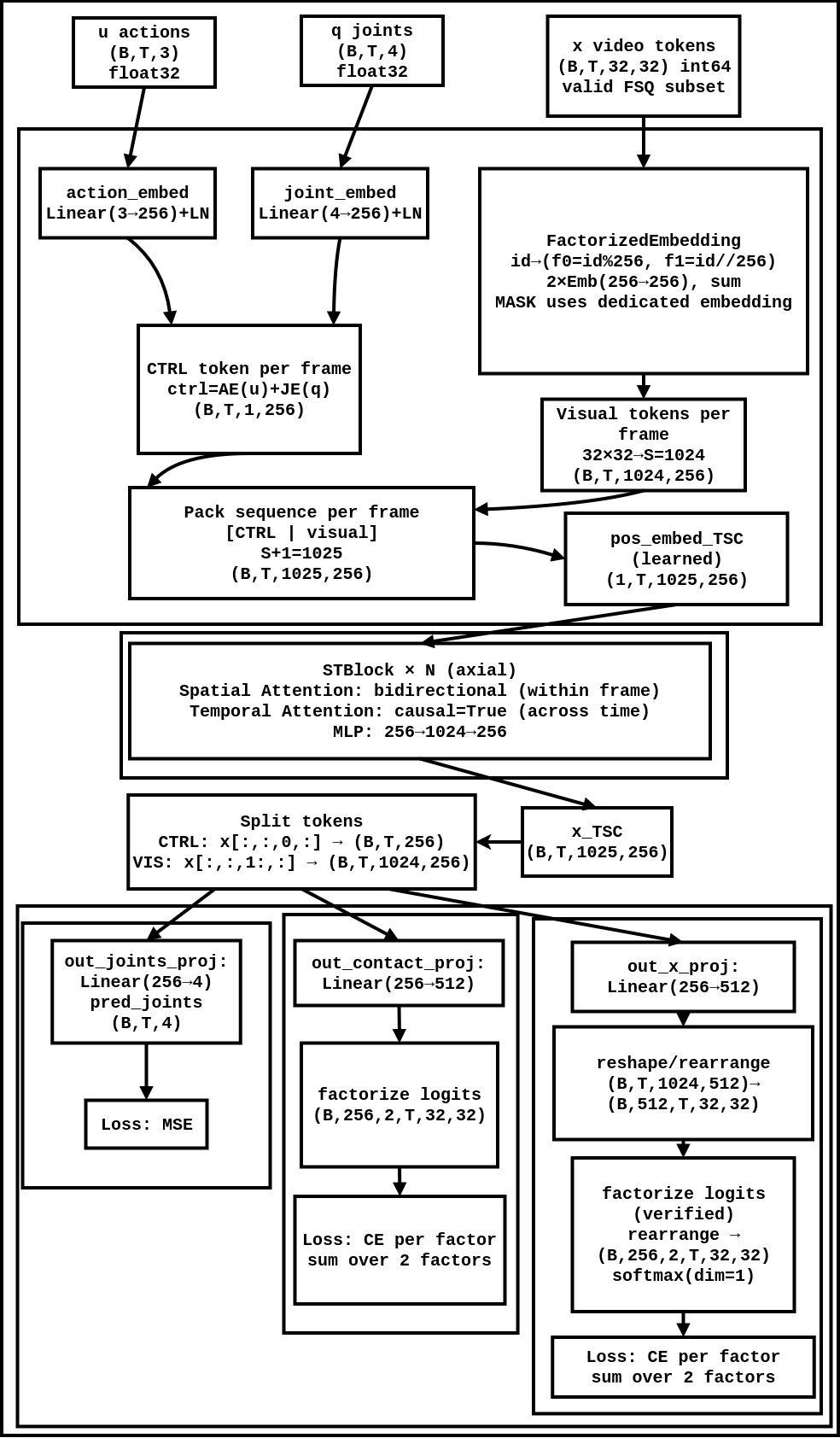}
  \caption{ST-Transformer architecture. Control tokens (action + joint embeddings) are concatenated with video tokens and processed by $N$ axial ST-Blocks. Three output heads: joint regression (MSE), contact tokens, and video tokens (factorized CE).}
  \label{fig:dynamics-model}
\end{figure}

Figure~\ref{fig:dynamics-model} details the transformer internals. The ST-Transformer decoder comprises $L = 24$ identical blocks, each applying factorized spatial and temporal attention followed by a feed-forward network. This factorization reduces complexity from $\mathcal{O}(T^2 S^2)$ for joint spatiotemporal attention to $\mathcal{O}(TS^2 + T^2S)$.

\paragraph{ST-Block Structure}

Each ST-Block contains three sub-layers applied in sequence: spatial self-attention across all $S+1$ tokens within each frame, temporal self-attention across all $T$ frames for each spatial position, and a position-wise feed-forward network.

\textit{Spatial Self-Attention.} Given input $\mathbf{X} \in \mathbb{R}^{B \times T \times (S+1) \times D}$, spatial attention operates independently on each frame. The input is first reshaped to $(BT, S+1, D)$:
\begin{equation}
    \mathbf{X}_{\text{spatial}} = \text{reshape}(\mathbf{X}, (BT, S+1, D))
\end{equation}

Standard multi-head self-attention is applied:
\begin{equation}
    \mathbf{X}_{\text{spatial}}' = \mathbf{X}_{\text{spatial}} + \text{SelfAttn}(\text{Norm}(\mathbf{X}_{\text{spatial}}))
\end{equation}

In spatial attention, the control token (at position 0) attends to and is attended by all video tokens (positions $1$ to $S$), enabling action-video interaction within each frame.

\textit{Temporal Self-Attention with Causal Masking.} For temporal attention, the tensor is reshaped to $(B(S+1), T, D)$:
\begin{equation}
    \mathbf{X}_{\text{temporal}} = \text{reshape}(\mathbf{X}_{\text{spatial}}', (B(S+1), T, D))
\end{equation}

Causal self-attention is applied with a lower-triangular attention mask:
\begin{equation}
    \mathbf{X}_{\text{temporal}}' = \mathbf{X}_{\text{temporal}} + \text{CausalSelfAttn}(\mathbf{X}_{\text{temporal}})
\end{equation}

The causal mask ensures that tokens at time $t$ can only attend to tokens at times $\leq t$:
\begin{equation}
    M_{ij} = \begin{cases}
        0 & \text{if } i \geq j \\
        -\infty & \text{otherwise}
    \end{cases}
\end{equation}

\paragraph{Self-Attention Mechanism}

The self-attention operation computes queries, keys, and values through linear projections:
\begin{align}
    \mathbf{Q} &= \mathbf{X} \mathbf{W}_Q \in \mathbb{R}^{N \times D} \\
    \mathbf{K} &= \mathbf{X} \mathbf{W}_K \in \mathbb{R}^{N \times D} \\
    \mathbf{V} &= \mathbf{X} \mathbf{W}_V \in \mathbb{R}^{N \times D}
\end{align}

Multi-head attention splits these into $N_h = 8$ heads with dimension $d_k = D / N_h = 32$:
\begin{equation}
    \text{head}_i = \text{softmax}\left(\frac{\mathbf{Q}_i \mathbf{K}_i^\top}{\sqrt{d_k}} + \mathbf{M}\right) \mathbf{V}_i
\end{equation}

The heads are concatenated and projected:
\begin{equation}
    \text{MultiHead}(\mathbf{X}) = \text{Concat}(\text{head}_1, \ldots, \text{head}_{N_h}) \mathbf{W}_O
\end{equation}

\textit{QK-Normalization.} Layer normalization is applied to queries and keys before computing attention scores:
\begin{align}
    \tilde{\mathbf{Q}}_i &= \text{LayerNorm}(\mathbf{Q}_i) \\
    \tilde{\mathbf{K}}_i &= \text{LayerNorm}(\mathbf{K}_i)
\end{align}

This stabilizes training, particularly for large models.

\textit{$\mu$P Scaling.} When using maximal update parameterization ($\mu$P), attention logits are scaled by $8/d_k$ instead of $1/\sqrt{d_k}$:
\begin{equation}
    \alpha = \frac{8}{d_k} \quad \text{($\mu$P)}
\end{equation}

This ensures consistent training dynamics across model widths.

\paragraph{Feed-Forward Network}

The feed-forward network consists of two linear transformations with GELU activation:
\begin{equation}
    \text{FFN}(\mathbf{x}) = \mathbf{W}_2 \cdot \text{Dropout}\left(\text{GELU}(\mathbf{W}_1 \mathbf{x} + \mathbf{b}_1)\right) + \mathbf{b}_2
\end{equation}

The hidden dimension is $D_{\text{ff}} = 4D = 1024$ (controlled by mlp\_ratio $= 4.0$).

The complete ST-Block forward pass is:
\begin{align}
    \mathbf{X}^{(1)} &= \mathbf{X} + \text{SpatialAttn}(\text{Norm}(\mathbf{X})) \\
    \mathbf{X}^{(2)} &= \mathbf{X}^{(1)} + \text{TemporalAttn}(\mathbf{X}^{(1)}) \\
    \mathbf{X}^{(3)} &= \mathbf{X}^{(2)} + \text{FFN}(\text{Norm}(\mathbf{X}^{(2)}))
\end{align}

\paragraph{Attention Pattern Analysis}

Spatial attention is bidirectional within each frame: the control token $\mathbf{C}_t$ attends to all video tokens $\{\mathbf{x}_{t,1}, \ldots, \mathbf{x}_{t,S}\}$, and each video token $\mathbf{x}_{t,s}$ attends to both the control token and all other video tokens in frame $t$. No causal mask is applied.

Temporal attention enforces causality across frames. Each position attends only to the same position at earlier times: the control token at time $t$ aggregates $\{\mathbf{C}_0, \ldots, \mathbf{C}_t\}$; the video token at position $s$ and time $t$ attends to $\{\mathbf{x}_{0,s}, \ldots, \mathbf{x}_{t,s}\}$. The causal mask prevents information flow from future frames.

\subsubsection{MaskGIT: Masked Generative Image Transformer}

\paragraph{Training Objective}

The model is trained using a masked language modeling objective adapted for visual tokens. During training, a subset of future frame tokens is masked, and the model predicts the original tokens.

\paragraph{Masking Strategy}

Given a sequence of $T$ frames, the first $T_h$ frames serve as context (history), and the remaining $T_f = T - T_h$ frames are candidates for masking.

For each training sample, a corruption strategy is applied:

\textit{Case 1: Standard MLM Masking.} With probability $1 - \rho_{\text{non-mlm}}$, the model applies per-frame masking. For each frame $t > 0$, a mask probability $p_t$ is sampled from a cosine schedule:
\begin{equation}
    p_t = \cos\left(\frac{\pi}{2} \cdot u\right), \quad u \sim \text{Uniform}(0, 1)
\end{equation}
Each spatial position $(h, w)$ is then independently masked with probability $p_t$:
\begin{equation}
    z_{t,h,w} \leftarrow \begin{cases}
        z_{\text{mask}} & \text{with probability } p_t \\
        z_{t,h,w} & \text{otherwise}
    \end{cases}
\end{equation}

\textit{Case 2: Autoregressive-Style Masking.} With probability $\rho_{\text{non-mlm}} = 0.5$, the model instead applies autoregressive-style masking. A boundary frame $t^* \sim \text{Uniform}\{T_h, \ldots, T-1\}$ is sampled; frames $\{0, \ldots, t^*-1\}$ remain fully unmasked while frames $\{t^*, \ldots, T-1\}$ undergo progressive corruption with increasing noise.

\paragraph{Random Token Corruption}

Prior to masking, tokens may be randomly corrupted:
\begin{equation}
    z_{t,h,w}^{(k)} \leftarrow \begin{cases}
        \text{Uniform}\{0, \ldots, V_f\!-\!1\} & \text{prob. } r \cdot u \\
        z_{t,h,w}^{(k)} & \text{otherwise}
    \end{cases}
\end{equation}
where $r = 0.2$ is the maximum corruption rate and $u \sim \text{Uniform}(0, 1)$ is sampled per batch.

\paragraph{Loss Computation}

\textit{Video Token Prediction Loss.} The model outputs logits for each factored vocabulary:
\begin{equation}
    \boldsymbol{\ell}_{t,s} = \mathbf{W}_{\text{out}} \mathbf{X}_{\text{video}, t, s} \in \mathbb{R}^{K \cdot V_f}
\end{equation}

These are reshaped to $(K, V_f)$ and cross-entropy loss is computed for each factored vocabulary:
\begin{equation}
    \mathcal{L}_{\text{video}} = \frac{1}{|\mathcal{M}|} \sum_{(t,s) \in \mathcal{M}} \sum_{k=1}^{K} \text{CE}(\boldsymbol{\ell}_{t,s}^{(k)}, z_{t,s}^{(k)})
\end{equation}
where $\mathcal{M}$ is the set of masked positions.

\textit{Contact Prediction Loss.} A separate output head predicts contact maps:
\begin{equation}
    \boldsymbol{\ell}^{\text{contact}}_{t,s} = \mathbf{W}_{\text{contact}} \mathbf{X}_{\text{video}, t, s}
\end{equation}

The contact loss uses the same factorized cross-entropy:
\begin{equation}
    \mathcal{L}_{\text{contact}} = \frac{1}{|\mathcal{M}|} \sum_{(t,s) \in \mathcal{M}} \sum_{k=1}^{K} \text{CE}(\boldsymbol{\ell}^{\text{contact}, (k)}_{t,s}, c_{t,s}^{(k)})
\end{equation}

\textit{Joint Angle Prediction Loss.} Joint angles are predicted from the control token embedding:
\begin{equation}
    \hat{\boldsymbol{\theta}}_t = \mathbf{W}_{\text{joint}} \mathbf{C}'_t \in \mathbb{R}^{N_j}
\end{equation}
where $\mathbf{C}'_t$ is the control token output from the transformer.

The joint loss is mean squared error over future frames:
\begin{equation}
    \mathcal{L}_{\text{joint}} = \frac{1}{T_f} \sum_{t=T_h}^{T-1} \|\hat{\boldsymbol{\theta}}_t - \boldsymbol{\theta}_t\|_2^2
\end{equation}

\textit{Total Loss.} The total training loss is:
\begin{equation}
    \mathcal{L} = \mathcal{L}_{\text{video}} + \lambda_{\text{contact}} \mathcal{L}_{\text{contact}} + \lambda_{\text{joint}} \mathcal{L}_{\text{joint}}
\end{equation}
with $\lambda_{\text{contact}} = 2.0$ and $\lambda_{\text{joint}} = 1.0$.

\paragraph{Inference: Iterative Decoding}

During inference, future frames are generated autoregressively, with each frame decoded using iterative parallel decoding.

\textit{Autoregressive Frame Generation.} For each future frame $t \in \{T_h, \ldots, T-1\}$, the model initializes all tokens as mask tokens, applies MaskGIT iterative decoding for $N_{\text{steps}}$ iterations, and then uses the predicted tokens as context for subsequent frames.

\textit{MaskGIT Iterative Decoding.} Within each frame, tokens are unmasked iteratively using a cosine schedule:

\begin{algorithm}[H]
\small
\caption{MaskGIT Decoding for Frame $t$}
\begin{algorithmic}[1]
\STATE Initialize: $\mathbf{z}_t^{(0)} \leftarrow \mathbf{z}_{\text{mask}}$ (all positions masked)
\FOR{$i = 0$ to $N_{\text{steps}} - 1$}
    \STATE Compute logits: $\boldsymbol{\ell} = f_\theta(\mathbf{z}_{<t}, \mathbf{z}_t^{(i)}, \mathbf{a}, \boldsymbol{\theta})$
    \STATE Sample tokens: $\hat{\mathbf{z}}_t \sim \text{Categorical}(\text{softmax}(\boldsymbol{\ell}/\tau))$
    \STATE Compute confidences: $c_s = \prod_{k} p(\hat{z}_{t,s}^{(k)} | \boldsymbol{\ell})$
    \STATE Compute mask ratio: $\gamma = \cos(\frac{\pi}{2} \cdot \frac{i+1}{N_{\text{steps}}})$
    \STATE Number to mask: $n = \lceil \gamma \cdot S \rceil$
    \STATE Select $n$ positions with lowest confidence to remain masked
    \STATE Update: $\mathbf{z}_t^{(i+1)}$, keeping high-confidence predictions
\ENDFOR
\STATE \textbf{return} $\mathbf{z}_t^{(N_{\text{steps}})}$
\end{algorithmic}
\end{algorithm}

\textit{Confidence Estimation.} For factorized tokens, confidence is the product of probabilities across vocabularies:
\begin{equation}
    c_s = \prod_{k=1}^{K} \max_{v} \text{softmax}(\boldsymbol{\ell}_s^{(k)})_v
\end{equation}

Two unmasking modes are implemented: \emph{greedy} selects tokens with highest confidence; \emph{random} samples uniformly and is applied during training to mitigate exposure bias.

\subsubsection{Video Decoder}

\paragraph{Decoder Architecture}

The decoder mirrors the encoder, mapping quantized latents to RGB images. Given quantized features $\hat{\mathbf{e}} \in \{-1, +1\}^{C_z \times h \times w}$, an initial $3 \times 3$ convolution projects to $C_{\text{base}} \cdot 4$ channels, followed by mid-block residual layers. Upsampling stages apply residual blocks with channel multipliers $(4, 2, 2, 1, 1)$ in reverse order. The output projection applies group normalization and Swish activation before a $3 \times 3$ convolution produces three RGB channels.

\paragraph{Upsampling Operation}

Spatial resolution is increased using depth-to-space (pixel shuffle) operations:
\begin{equation}
    \text{Upsample}(\mathbf{h}) = \text{DepthToSpace}(\text{Conv}_{3\times 3}(\mathbf{h}; 4C), \text{block\_size}=2)
\end{equation}

This rearranges a tensor of shape $(B, 4C, H, W)$ to $(B, C, 2H, 2W)$.

\paragraph{Token-to-Latent Mapping}

Given predicted token indices $z \in \{0, \ldots, V\!-\!1\}$, the decoder recovers the continuous latent representation via the inverse FSQ de-quantization operation. This produces a tensor $\mathbf{e} \in \mathbb{R}^{C \times h \times w}$ with $C = 6$ channels, which is then passed to the decoder network for image reconstruction. The inverse Haar wavelet transform is applied during decoding to upsample the spatial resolution by a factor of 4.

\paragraph{Factorization Strategy} For computational efficiency during training and inference, token indices are factorized into $K = 2$ factors, each with vocabulary size $V_f = 256$, such that $V_f^K = 65{,}536 \approx V$. This reduces the output logit dimension from $V$ to $K \cdot V_f$ and enables parallel prediction across factors.

\subsubsection{Model Output Specification}

\paragraph{Forward Pass Outputs}

During training, the forward pass returns the total loss $\mathcal{L}$ and its components $\mathcal{L}_{\text{video}}$, $\mathcal{L}_{\text{contact}}$, $\mathcal{L}_{\text{joint}}$, along with token-level accuracies on masked positions. The full logit tensor of shape $(B, K \cdot V_f, T, H, W)$ is retained for downstream analysis.

\paragraph{Generation Outputs}

At inference, \texttt{generate()} returns predicted video token indices of shape $(B, T \cdot H \cdot W)$. Optional outputs include contact map tokens $(B, T_f, H, W)$, predicted joint angles $(B, T_f, N_j)$, and factored logits for uncertainty quantification.

\subsubsection{Model Configuration}

\paragraph{Default Hyperparameters}

The model uses $L = 24$ transformer layers with hidden dimension $D = 256$, $N_h = 8$ attention heads (head dimension $d_k = 32$), FFN hidden dimension $D_{\text{ff}} = 1024$, temporal sequence length $T = 16$, and spatial sequence length $S = 1024$ ($32 \times 32$ tokens per frame). The image vocabulary size is $V = 65536$, factorized into $K = 2$ sub-vocabularies of size $V_f = 256$ each. Joint angle inputs have $N_j = 4$ channels.

\paragraph{Computational Complexity}

Per ST-Block, spatial attention requires $\mathcal{O}(BT(S+1)^2 D)$ operations while temporal attention requires $\mathcal{O}(B(S+1)T^2 D)$ operations. For $L$ layers, the total complexity is:
\begin{equation}
    \mathcal{O}(L \cdot BTS(S + T)D)
\end{equation}

With default parameters ($T=16$, $S=1024$, $D=256$, $L=24$), this is approximately $\mathcal{O}(10^{11})$ operations per forward pass for batch size 1.

\textit{Parameter Count.} The model contains approximately:
\begin{equation}
    \text{Params} \approx L \cdot (12D^2 + 8D^2) + V_f \cdot K \cdot D + T \cdot (S+1) \cdot D
\end{equation}
yielding roughly 30-50M parameters depending on configuration.

\subsection{Online World Model Evaluation in the Planning Loop}

\subsubsection{Coupling ChronoDreamer in Eval Simulation}

During each evaluation cycle, ChronoDreamer aggregates a short history of camera frames, robot states, and executed actions, then feeds these sequences into the world model to predict next-step RGB observations and contact maps. The predicted futures, alongside the tiled history, are passed to an LLM judge that reasons about collision risk.

The planner samples candidate actions and submits them to this LLM gate synchronously each world-model period; any action flagged as unsafe is resampled, up to a fixed maximum of three attempts. If all sampled actions are rejected, the controller falls back to a conservative retreat---commanding an upward ($-Z$) motion---to clear the workspace before continuing.

\subsubsection{LLM Collision Judge}

The following prompt template elicits collision assessments from the LLM:

\vspace{2mm}
\noindent
\colorbox{yellow!15}{%
\begin{minipage}{0.98\columnwidth}
\scriptsize\ttfamily
\setlength{\parindent}{0pt}
\setlength{\parskip}{1pt}

LLM\_PROMPT\_TEMPLATE = """Do you think there is a collision happening between the manipulator and any of the objects on the table? The camera is mounted on the arm of the manipulator. You are a strict collision verifier.

You will be shown:\\
(1) History RGB frames (prompt)\\
(2) Predicted future RGB frames (generated)\\
(3) Predicted contact map (generated) -- WARNING: this is noisy and often has false positives.

TASK\\
Decide if a collision happens between the manipulator/gripper and ANY table object.

DEFINITION\\
Output collision\_likely=true ONLY if BOTH are true:\\
(A) In the predicted future RGB frames, the gripper/manipulator makes physically plausible contact with an object (not just overlaps in 2D).\\
(B) At least one of the following is visible in RGB:\\
\quad (B1) The object moves/rotates/shifts relative to the table (displacement) compared to its pose in the last prompt frame.\\
\quad (B2) The object deforms/deflects/compresses at the contact point (common for chains/FEA/deformables), even if its base pose does not translate.\\
\quad (B3) There is sustained pushing/pressing contact across multiple future frames with no visible gap, even if the object is constrained/heavy and does not translate.

IMPORTANT RULES (to reduce false positives)\\
- Do NOT claim collision based only on the contact map. The contact map can be wrong.\\
- If the gripper crosses over an object in the image in 2D but there is a visible gap/clearance (no sustained contact), this is NOT a collision.\\
- If you cannot clearly see contact evidence in the predicted RGB frames (displacement, deformation/deflection, or sustained pressing contact), set collision\_likely=false.\\
- Lack of object translation does NOT imply "no collision" when the object is constrained/heavy or deformable. In those cases, look for deformation/deflection or sustained pressing contact.\\
- If the yellow table seems becoming smaller, this likely means the arm/camera is lifting upward; lifting makes collision less likely. Prefer collision\_likely=false unless there is clear contact evidence.\\
- Only give high confidence ($>$= 0.85) when contact evidence is obvious and sustained across multiple future frames.\\
- If evidence is ambiguous or blurred, set collision\_likely=false and confidence $<$= 0.5.

FEW-SHOT EXAMPLES (read carefully)\\
Example 1 (NO collision: high clearance pass-over)\\
\quad Predicted frames show the gripper moving "over" a block in 2D, but the block's pose does not change relative to the table. Even if the contact map lights up near the block, answer:\\
\quad \{"collision\_likely": false, "confidence": 0.3, "first\_collision\_frame": 0, "explanation": "Gripper passes above/near object; no visible displacement in RGB."\}

Example 2 (NO collision: apparent motion from camera ego-motion)\\
\quad Camera moves with the arm; objects may appear to shift slightly due to viewpoint changes. If the object remains fixed relative to table edges/markers and there is no clear push/impact, answer:\\
\quad \{"collision\_likely": false, "confidence": 0.4, "first\_collision\_frame": 0, "explanation": "No clear object displacement; apparent changes likely from camera motion."\}

Example 3 (YES collision: clear push)\\
\quad Gripper contacts a block and the block translates/rotates in a sustained way across multiple future frames. Answer:\\
\quad \{"collision\_likely": true, "confidence": 0.9, "first\_collision\_frame": 1, "explanation": "Contact with sustained object displacement visible in RGB."\}

Example 4 (YES collision: constrained/heavy/deformable object)\\
\quad The gripper presses into a chain/beam/deformable object and you can see bending/compression/deflection at the contact point, even if the object does not translate. Answer:\\
\quad \{"collision\_likely": true, "confidence": 0.85, "first\_collision\_frame": 1, "explanation": "Sustained pressing contact with visible deformation/deflection in RGB."\}

Inputs:\\
- Image 1: history RGB frames (prompt)\\
- Image 2: predicted future RGB frames (generated)\\
- Image 3: predicted future contact map (generated)

Return JSON only with the following schema:\\
\{\\
\quad "collision\_likely": true/false,\\
\quad "confidence": 0.0-1.0,\\
\quad "first\_collision\_frame": 0-7,\\
\quad "explanation": "..."\\
\}\\
"""

\end{minipage}%
}
\vspace{2mm}

\section{Experiments}
\label{sec:experiments}

We evaluate on \emph{DreamerBench}, a simulator-generated dataset targeting frictional, contact-rich tabletop dynamics~\cite{zhou2025dreamerbench}.
It complements DMControl~\cite{tassa2018dmcontrol}, ALE~\cite{bellemare2013ale}, D4RL~\cite{fu2020d4rl}, RLBench~\cite{james2019rlbench}, Open X-Embodiment~\cite{openxembodiment2023}, DrivingDojo~\cite{wang2024drivingdojo}, and WorldPrediction~\cite{chen2025worldprediction} by exposing dense contact supervision.

\subsection{Trajectory Generation and State Logging}

Trajectories are generated using Project Chrono, configured as a
multi-physics engine with rigid and (optionally) flexible bodies,
Coulomb friction, and penalty-based or constraint-based contact
handling. At a fixed simulation time step $\Delta t$, the underlying
continuous-time dynamics can be interpreted as a hybrid ODE/DAE system
with a non-smooth right-hand side; DreamerBench records a discretized
version of these dynamics suited for sequence modeling.

At each simulation step $t$, a state tuple
\begin{equation}
  s_t = \bigl(
    o_t^{\text{rgb}},
    o_t^{\text{contact}},
    q_t, \dot{q}_t,
    f_t^{\text{contact}},
    \mu_t,
    c_t
  \bigr)
\end{equation}
is logged, together with a low-level action $a_t$ and, optionally, a task reward
$r_t$, where:
\begin{itemize}
  \item $o_t^{\text{rgb}}$ collects RGB images from one or more cameras
  (e.g., ego and side views), at a resolution suitable for
  vector-quantized autoencoders.

  \item $o_t^{\text{contact}}$ is a ``contact splat'' image: a
  workspace-aligned grid that encodes the spatial footprint and
  magnitude of contact (e.g., approximated contact pressure).

  \item $q_t,\dot{q}_t$ are robot joint positions and velocities,
  forming a proprioceptive observation channel.

  \item $f_t^{\text{contact}}$ aggregates per-contact $3$D normal and
  tangential forces (either per contact point or grouped per body),
  corresponding to impulses in non-smooth
  mechanics discretized over $\Delta t$.

  \item $\mu_t$ denotes scalar or per-body friction coefficients used
  by the contact model (e.g., static and kinetic coefficients
  $(\mu_{\mathrm{s}},\mu_{\mathrm{k}})$), enabling variation
  of friction regimes.

  \item $c_t$ is a discrete contact-mode flag (no contact, sticking,
  sliding, separating), which can be interpreted as a latent
  discrete state in a hybrid latent variable model.
\end{itemize}

Each episode (trajectory) is represented as
\begin{equation}
  \tau = (s_0, a_0, r_0, s_1, a_1, r_1, \dots, s_T),
\end{equation}
with horizon $T$ chosen so that objects undergo multiple contact mode
transitions (impact, stick--slip, rolling, and settling). This setting is suitable for evaluating long-horizon rollout
stability of world models under hybrid dynamics. Episodes last 300--600 steps to capture multiple stick--slip transitions.

\subsection{Simulation Scenarios}

Each scenario---\texttt{flashlight-box}, \texttt{flashlight-coca}, \texttt{waterbottle-coca}, and \texttt{fea-flashlight}---provides synchronized RGB views, contact splat images, proprioception, low-level actions, and physics annotations (forces, friction coefficients, mode flags), and vector-quantized latents are pre-computed so that token-based models avoid repeated encoding.
The domains are deliberately non-smooth: transition kernels switch whenever contacts appear or disappear, friction regimes change, or rolling transitions to sticking. Table~\ref{tab:dreamerbench_scenarios} illustrates the four scenarios with example RGB and contact-splat images.

\begin{table*}[t]
  \centering
  \small
  \caption{Simulation scenarios in DreamerBench with example RGB and contact-splat images.}
  \label{tab:dreamerbench_scenarios}
  \setlength{\tabcolsep}{3pt}
  \renewcommand{\arraystretch}{1.05}
  \begin{tabularx}{\textwidth}{L{0.18\textwidth} Y C{0.14\textwidth} C{0.14\textwidth} C{0.14\textwidth}}
    \toprule
    Scenario & Description & Ego camera & Side camera 0 & Contact-splat image \\
    \midrule
    \texttt{flashlight box} &
    Tool interacting with a box on a planar surface with sliding and pushing. &
    \includegraphics[width=0.9\linewidth]{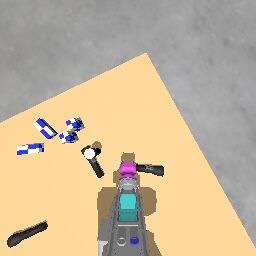} &
    \includegraphics[width=0.9\linewidth]{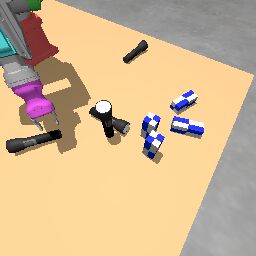} &
    \includegraphics[width=0.9\linewidth]{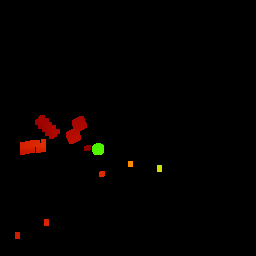} \\
    \addlinespace[3pt]
    \texttt{flashlight coca} &
    Tool interacting with a tall cylindrical container with sliding and rolling contact. &
    \includegraphics[width=0.9\linewidth]{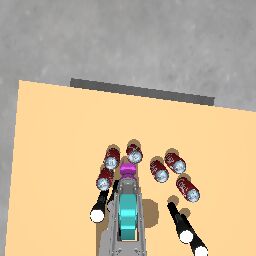} &
    \includegraphics[width=0.9\linewidth]{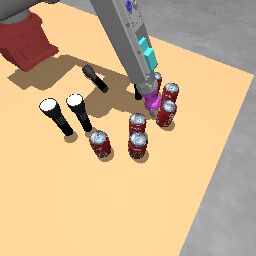} &
    \includegraphics[width=0.9\linewidth]{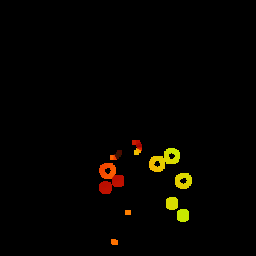} \\
    \addlinespace[3pt]
    \texttt{waterbottle coca} &
    Bottle--cylinder interaction with combined sliding and rolling along the table. &
    \includegraphics[width=0.9\linewidth]{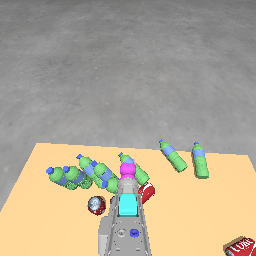} &
    \includegraphics[width=0.9\linewidth]{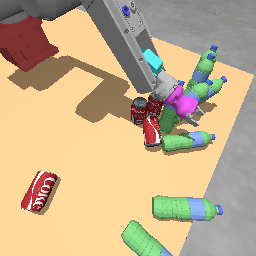} &
    \includegraphics[width=0.9\linewidth]{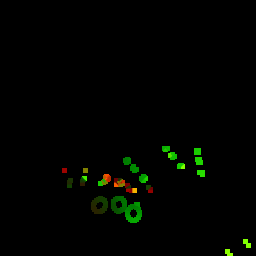} \\
    \addlinespace[3pt]
    \texttt{fea flashlight} &
    Four tetrahedron-defined FEA beams and two standing flashlights represented as rigid bodies. &
    \includegraphics[width=0.9\linewidth]{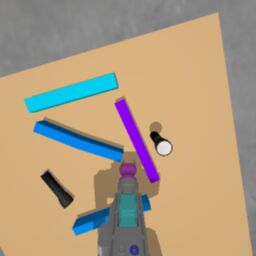} &
    \includegraphics[width=0.9\linewidth]{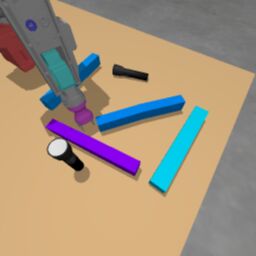} &
    \includegraphics[width=0.9\linewidth]{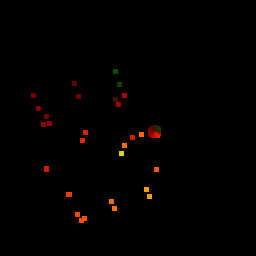} \\
    \bottomrule
  \end{tabularx}
\end{table*}

\paragraph{File organization and pre-computed latents.}
On disk, DreamerBench is organized into scenario-specific archives (e.g.,
\texttt{flashlight\_box\_00.zip}), each containing multiple episodes and
auxiliary files:

\begin{itemize}
  \item arrays for RGB and contact-splat images, proprioception, actions,
  and physics annotations;

  \item pre-computed discrete visual tokens
  $\mathbf{z}_t \in \{0,\dots,K-1\}^S$ for RGB and contact streams,
  obtained from a vector-quantized autoencoder and stored in binary
  files (e.g., \texttt{video.bin}, \texttt{contact\_splat.bin});

  \item metadata (JSON) describing token grid size $S$, codebook size
  $K$, frame count, frame rate, and segmentation into episodes.
\end{itemize}

This layout is compatible with common training stacks (PyTorch, JAX)
that either stream episodes as variable-length sequences or treat the
dataset as a concatenated sequence with associated \emph{segment
IDs}. ``Contact splat'' images are generated with the encoder in \Cref{subsec:dreamerbench:contact-encoding} and stored alongside tokenized latents $\mathbf{z}_t$ plus metadata for streaming.

\subsection{Stochastic Joystick Excitation via an Ornstein--Uhlenbeck Process}
\label{subsec:dreamerbench:ou}

To generate temporally correlated yet nontrivial end-effector motion for data collection, and to excite rich contact behaviors, the joystick command channel in \emph{DreamerBench} is driven by an Ornstein--Uhlenbeck (OU) process. The OU process is a stationary Gaussian Markov process originally introduced as a mean-reverting model for Brownian velocity fluctuations~\cite{uhlenbeck1930brownian}. In continuous time, a $d$-dimensional OU process $X_t \in \mathbb{R}^d$ satisfies the Langevin-type stochastic differential equation
\begin{equation}
  \mathrm{d}X_t
  = \theta (\mu - X_t)\,\mathrm{d}t
  + \sigma\,\mathrm{d}W_t,
  \label{eq:ou_sde}
\end{equation}
where $\theta>0$ is the mean-reversion rate, $\mu \in \mathbb{R}^d$ is the long-term mean, $\sigma>0$ is the diffusion scale, and $W_t$ is $d$-dimensional standard Brownian motion. The process has an exponential autocorrelation function with correlation time $\tau_c = 1/\theta$ and a Gaussian stationary distribution; efficient numerical schemes for OU trajectories and related functionals are standard in the SDE literature~\cite{gillespie1996ou}.

OU noise is widely used in continuous-control reinforcement learning as a model of colored action noise, in particular to drive exploration in physical systems with inertia without introducing unrealistically high-frequency jitter~\cite{lillicrap2015ddpg,hollenstein2022actionnoise}. Here, the same construction is used not for exploration during learning, but to synthesize joystick-like excitation that repeatedly engages contact, sliding, and rolling during data collection. We instantiate a three-dimensional OU process
\begin{equation}
  X_t = \bigl( X_t^{(x)}, X_t^{(y)}, X_t^{(z)} \bigr)^\top
\end{equation}
representing a latent control signal over Cartesian directions $(x,y,z)$ in the workspace of the robot arm.

The implementation mirrors an Euler--Maruyama discretization of~\eqref{eq:ou_sde} with time step $\Delta t$,
\begin{equation}
  X_{k+1}
  = X_k
    + \theta (\mu - X_k)\,\Delta t
    + \sigma \sqrt{\Delta t}\,\xi_k,
  \qquad
  \xi_k \sim \mathcal{N}(0, I_3),
  \label{eq:ou_discrete}
\end{equation}
which is encoded in the \texttt{OrnsteinUhlenbeckProcess.sample()} method. The parameters used in the script correspond to relatively large $\theta$ and $\sigma$, yielding high-variance but strongly time-correlated excitation, broadly consistent with common practice in continuous-control benchmarks~\cite{lillicrap2015ddpg,hollenstein2022actionnoise}.

After each update~\eqref{eq:ou_discrete}, the process state is post-processed before being interpreted as a joystick command. First, a minimum-magnitude constraint is applied: letting $m_k = \|X_k\|_2$ and denoting a threshold $m_{\min}>0$, the state is rescaled according to
\begin{equation}
  X_k \leftarrow \begin{cases}
    (m_{\min}/m_k) X_k, & 0 < m_k < m_{\min}, \\[4pt]
    v\,m_{\min}, & m_k = 0,
  \end{cases}
\end{equation}
where $v$ is a random unit vector. This step enforces a lower bound on the norm of $X_k$ while preserving the current direction whenever possible, avoiding extended intervals of near-zero actuation.

Next, each component is passed through a deadzone and mapped to a Cartesian increment. For component $X_k^{(i)}$, a threshold $\varepsilon>0$ is applied:
\begin{equation}
  X_k^{(i)} \leftarrow \begin{cases}
    0, & |X_k^{(i)}| < \varepsilon, \\
    X_k^{(i)}, & \text{otherwise,}
  \end{cases}
\end{equation}
and the resulting vector updates the end-effector position:
\begin{equation}
  p_{k+1} = \Pi_{\mathcal{P}}\bigl(p_k + v_{\text{scale}}\,X_k\,\Delta t\bigr),
\end{equation}
where $v_{\text{scale}}$ is a scalar step size and $\Pi_{\mathcal{P}}$ denotes component-wise clipping to a predefined axis-aligned workspace box $\mathcal{P}$. The updated target position $p_{k+1}$ is then passed to an inverse kinematics solver, which produces joint angles applied to the robot at the control rate.

Overall, the joystick control law in \emph{DreamerBench} can be viewed as an OU-driven colored-noise excitation filtered through a norm constraint, a deadzone nonlinearity, and a constrained kinematic integrator. This construction yields temporally coherent, reproducible stochastic motion that systematically excites contact-rich interactions while avoiding degenerate trajectories with negligible motion, and it admits straightforward ablations (e.g., modifying $\tau_c$ or replacing OU noise with i.i.d.\ Gaussian noise) for future studies. This colored-noise excitation consistently drives sliding, rolling, and intermittent sticking without degenerating into near-static trajectories.

\section{Results}
\label{sec:results}

\subsection{Ablations Studies}
We are actively working on ablation studies to understand the scaling and the of the performance of different network sizes with number of tasks trained on. We will report the results here soon. Due to very limited compute resources, we have not been able to run these experiments yet.

\subsection{Qualitative Analysis}

A qualitative analysis has been conducted by human to gain understanding on the model performance after training for 4 epochs using the following parameters.

\begin{figure}[t]
	\centering
	\begin{subfigure}[b]{0.49\textwidth}
		\includegraphics[width=\textwidth]{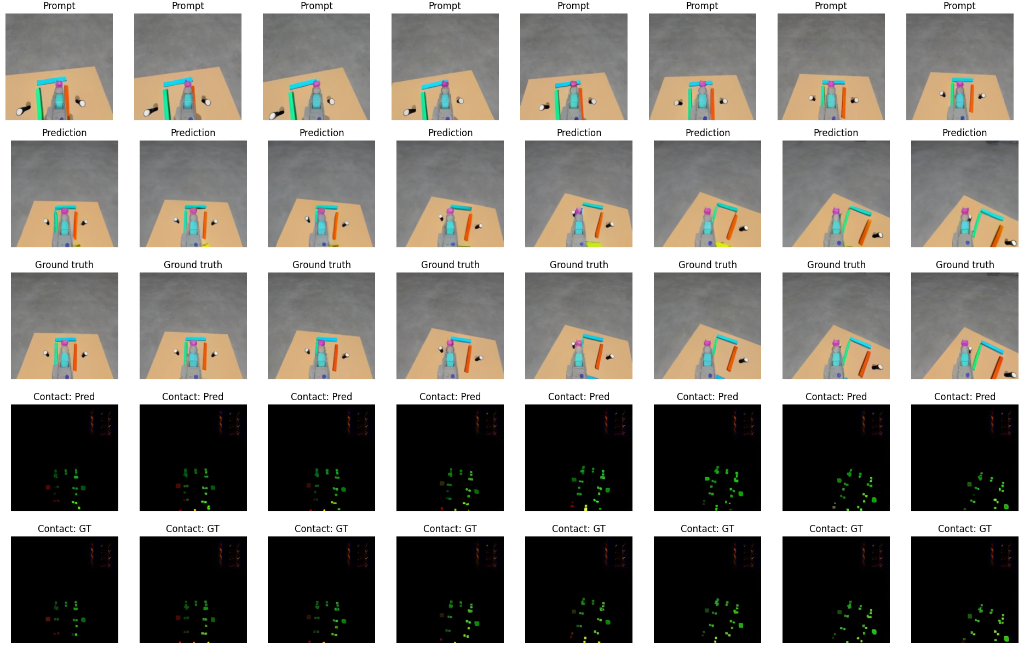}
		\caption{End-effector moving left.}
		\label{fig:spatial_1}
	\end{subfigure}
	\\[2pt]
	\begin{subfigure}[b]{0.49\textwidth}
		\includegraphics[width=\textwidth]{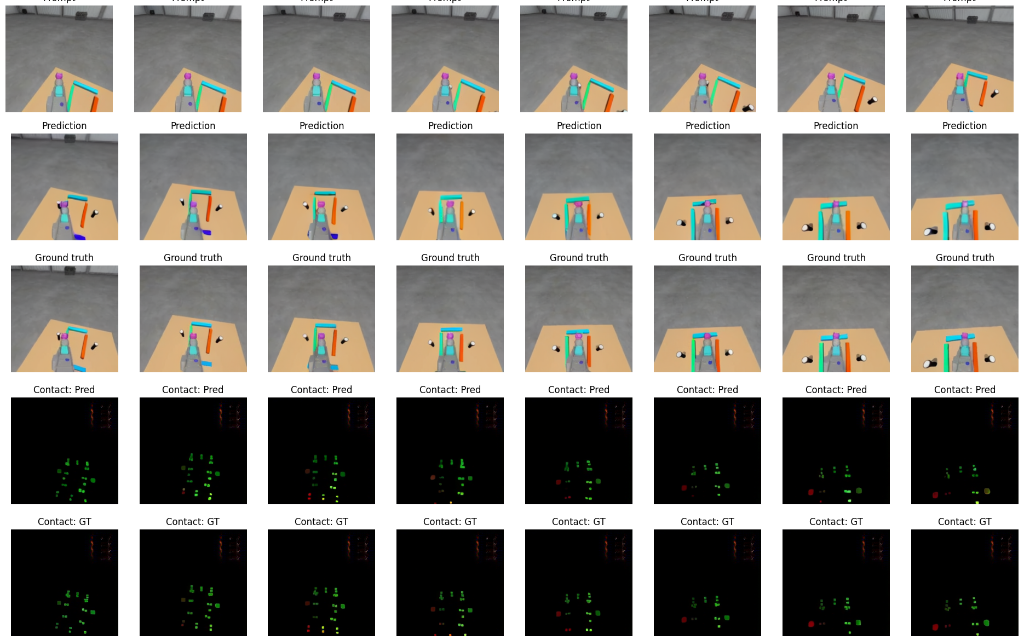}
		\caption{End-effector moving right, hovering above FEA beam.}
		\label{fig:spatial_2}
	\end{subfigure}
	\\[2pt]
	\begin{subfigure}[b]{0.49\textwidth}
		\includegraphics[width=\textwidth]{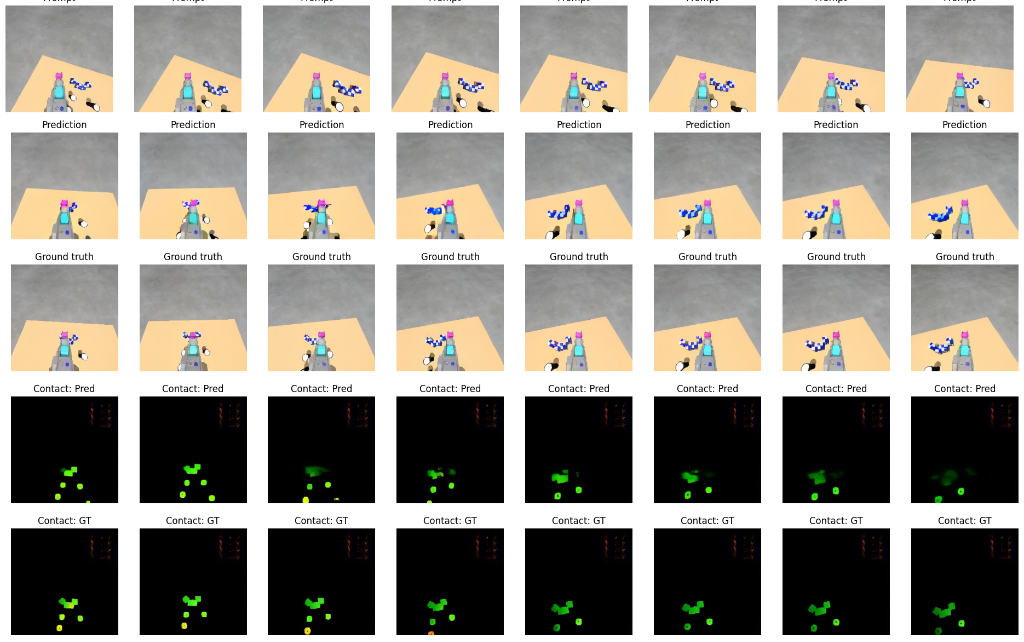}
		\caption{End-effector moving right.}
		\label{fig:spatial_3}
	\end{subfigure}
	\caption{Spatial coherence in non-contact scenarios (\texttt{fea\_flashlight}). Row~1: prompt; Rows~2--3: predicted RGB; Rows~4--5: predicted contact splat. Time progresses left to right.}
	\label{fig:spatial_coherence}
\end{figure}

We have noticed the spatial and object coherence, the objects' relative positions on the table, can be preserved easily by the model, across multiple scenarios, when contact has not happened. Examples have been provided in Figure~\ref{fig:spatial_coherence}. The spatial coherence is a combination of camera rendering and positioning, lower level inverse kinematic controller, and multibody dynamics physics.

\begin{figure}[t]
	\centering
	\begin{subfigure}[b]{0.49\textwidth}
		\includegraphics[width=\textwidth]{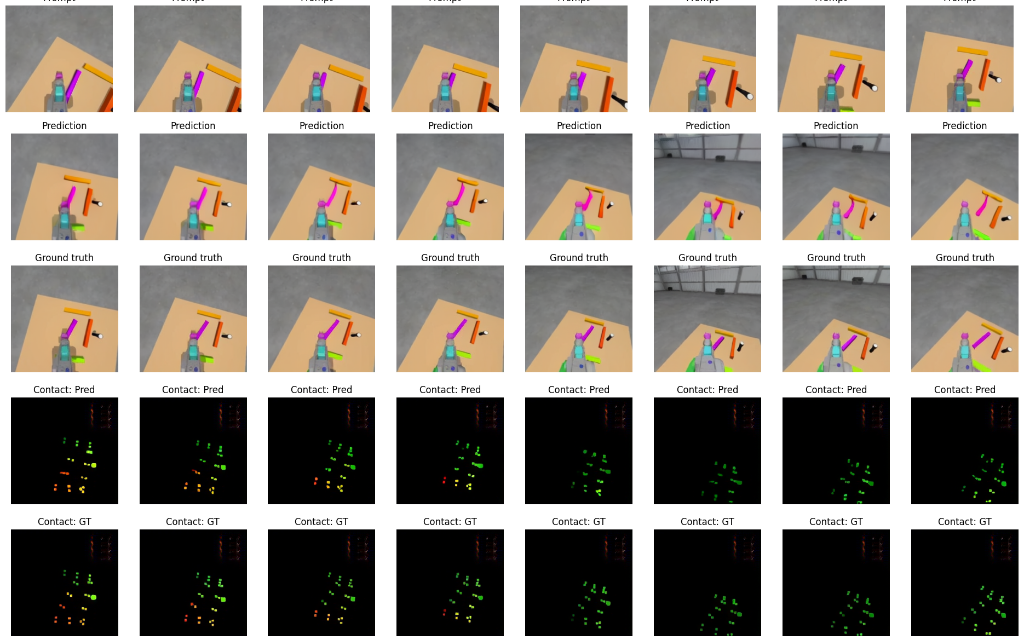}
		\caption{\texttt{fea\_flashlight}: gripper collides with deformable beam.}
		\label{fig:contact_1}
	\end{subfigure}
	\\[2pt]
	\begin{subfigure}[b]{0.49\textwidth}
		\includegraphics[width=\textwidth]{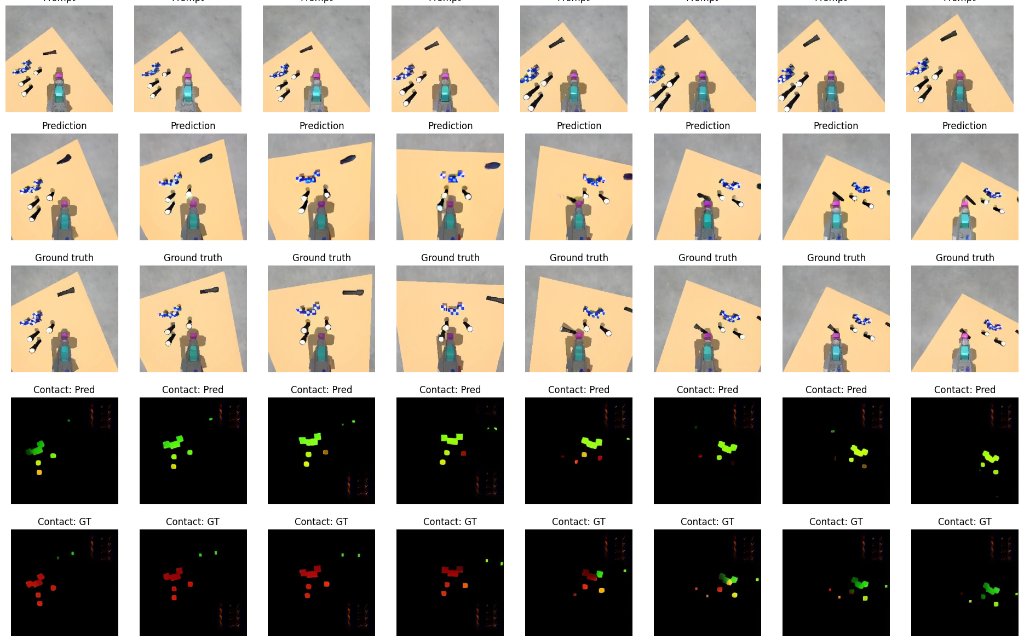}
		\caption{\texttt{flashlight\_box}: gripper pushes flashlight leftward.}
		\label{fig:contact_2}
	\end{subfigure}
	\caption{Predictions during contact events. Layout as in Fig.~\ref{fig:spatial_coherence}. Post-contact frames exhibit increased blurriness.}
	\label{fig:contact_examples}
\end{figure}

Figures~\ref{fig:contact_1} and~\ref{fig:contact_2} show cases in which a contact happens between the end-effector and an object on the table. The model is able to predict the contact event and generate corresponding contact splat frames. However, the predicted video and camera frames tend to show artifacts and blurriness after contact happens, and certain spatial coherence is lost. This behavior is quite consistent cross multiple scenarios, tasks, and given prompts. This might be an indication that the current DreamerBench dataset does not contain enought contact-rich data, as indeed due to the natural of action sampling and simulation setup, contact events take a relatively small portion of the entire dataset.

\section{Discussion}
Qualitatively, the trained world model is capable of deducing possible collision in most cases. We have noticed prompt engineering is the key to guide llm to generate correct collision flags. An example correct reasoning case is shown below, Gemma-3-27b-it from Google Gemini family is used for the example rasoning:

\begin{figure}[t]
	\centering
	\begin{subfigure}[b]{0.49\textwidth}
		\includegraphics[width=\textwidth]{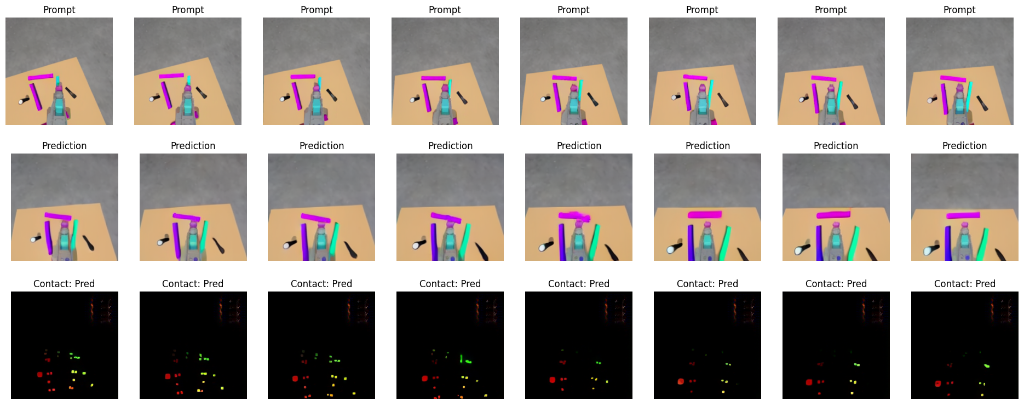}
		\caption{Collision detected (conf.\ 0.95): manipulator pushes object toward table edge.}
		\label{fig:reasoning-attempt0}
	\end{subfigure}
	\\[2pt]
	\begin{subfigure}[b]{0.49\textwidth}
		\includegraphics[width=\textwidth]{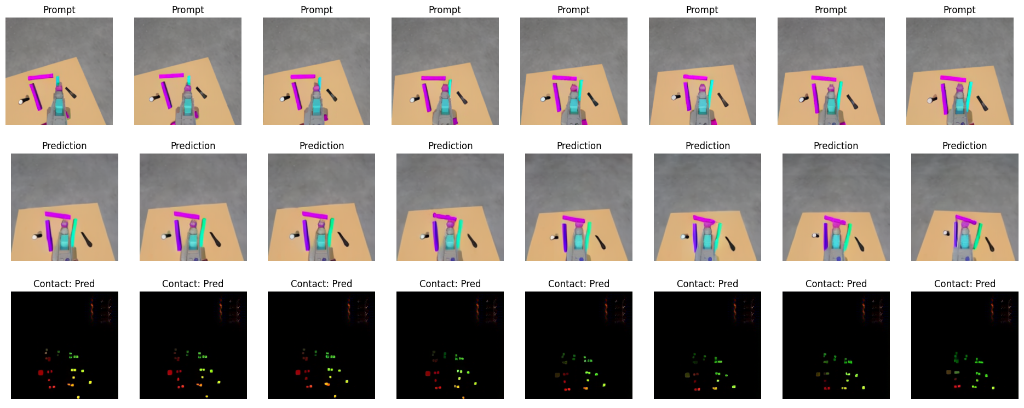}
		\caption{No collision (conf.\ 0.3): manipulator lifts away from table.}
		\label{fig:reasoning-attempt1}
	\end{subfigure}
	\caption{Consecutive action samples at the same timestep. LLM judge rejects attempt~0, accepts attempt~1.}
	\label{fig:reasoning-attempts}
\end{figure}

Figure \ref{fig:reasoning-attempt0} and Figure \ref{fig:reasoning-attempt1} show two consecutive action sampling at the same time stamp, this example case shows the action sampling associated with attempt 0, in Figure \ref{fig:reasoning-attempt0}, with an llm reasoning giving a collision confidence of 0.95, would result in a collision. Then, the system moved on to perform a resampling and the world model predicted the rollout in attempt 1, with the generated frame and contact, the world model predicted a collision confidence of 0.3, decided the action sampling is categorized as safe. In this example case, in Figure \ref{fig:reasoning-attempt0}, the world model correctly identified a collision case where the manipulator pushes the deformable object towards the edge of the table. The generated frames depict the movement of the deformable object being pushed by the manipulator. In Figure \ref{fig:reasoning-attempt1}, the world model recognized a non-collision case where the manipulator lifts up and moves away from both the table and the deformable object. The generated frames effectively illustrate this movement.

The llm reasonings for both of these attempts are shown below:

For attempt 0 (Collision Case) - Response:

\vspace{2mm}
\noindent
\colorbox{yellow!15}{%
\begin{minipage}{0.98\columnwidth}
\scriptsize\ttfamily
\setlength{\parindent}{0pt}
\setlength{\parskip}{1pt}

Attempt 0 (Collision Case) - Gemini Response:\\
\{\\
\quad "collision\_likely": true,\\
\quad "confidence": 0.95,\\
\quad "first\_collision\_frame": 4,\\
\quad "explanation": "The gripper visibly contacts the small cylindrical object and pushes it to the right across multiple future frames. Clear displacement is visible in the RGB images."\\
\}\\

Model: gemma-3-27b-it\\
Prompt Token Count: 760 (TEXT) + 774 (IMAGE) = 1534 total\\
Total Token Count: 1534\\
Response ID: r8FEaceAGsG-\_uMPrtOugQc

\end{minipage}%
}
\vspace{2mm}

For attempt 1 (Non-Collision Case) - Response:

\vspace{2mm}
\noindent
\colorbox{yellow!15}{%
\begin{minipage}{0.98\columnwidth}
\scriptsize\ttfamily
\setlength{\parindent}{0pt}
\setlength{\parskip}{1pt}

Attempt 1 (Non-Collision Case) - Gemini Response:\\
\{\\
\quad "collision\_likely": false,\\
\quad "confidence": 0.3,\\
\quad "first\_collision\_frame": 0,\\
\quad "explanation": "Gripper approaches object but no visible displacement of the object is observed in the predicted RGB frames. Apparent changes could be due to camera motion."\\
\}\\

Model: gemma-3-27b-it\\
Prompt Token Count: 760 (TEXT) + 774 (IMAGE) = 1534 total\\
Total Token Count: 1534\\
Response ID: ysFEaaOaL7iv\_uMP2PX-yAg

\end{minipage}%
}
\vspace{2mm}

\section{Future Work}
The current version of ChronoDreamer shows good generalization across multiple different scenarios and tasks. So far, only random excitation has been used for simulation evaluation, we reported qualitative success to prevent potential collision in a random OU action sampling setting. In the future, we plan to integrate ChronoDreamer with a model predictive controller (MPC) or VLA to perform closed-loop control for specific tasks, such as reaching a target position while avoiding collisions with objects. Based on the current results using a small network, world model has already shown its capability for hazard prevention and can potentially increase controller success rate. 

\clearpage
\bibliographystyle{asmems4}
\bibliography{BibFiles/refsGraphics,BibFiles/refsSensors,BibFiles/refsAutonomousVehicles,BibFiles/refsChronoSpecific,BibFiles/refsDEM,BibFiles/refsFSI,BibFiles/refsMBS,BibFiles/refsRobotics,BibFiles/refsSBELspecific,BibFiles/refsTerramech,BibFiles/refsCompSci,BibFiles/refsNumericalIntegr,BibFiles/refsMLPhysics,BibFiles/refsSurfaceTension,BibFiles/refsStatsML,BibFiles/refsOddsEnds,BibFiles/refsML-AI}


\end{document}